# Revisiting Salient Object Detection: Simultaneous Detection, Ranking, and Subitizing of Multiple Salient Objects

Md Amirul Islam*  
University of Manitoba  
amirul@cs.umanitoba.ca

Mahmoud Kalash*  
University of Manitoba  
kalashm@cs.umanitoba.ca

Neil D. B. Bruce  
Ryerson University  
bruce@ryerson.ca

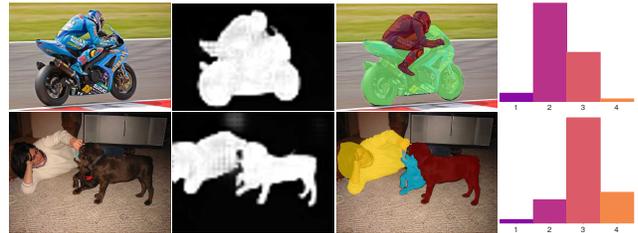

Figure 1. We present a solution in the form of a deep neural network to detect salient objects, consider the relative ranking of salience of these objects, and predict the total number of salient objects. Left to right: input image, detected salient regions, rank order of salient objects, confidence score for salient object count (subitizing). Colors indicate the rank order of different salient object instances.

## Abstract

*Salient object detection is a problem that has been considered in detail and many solutions proposed. In this paper, we argue that work to date has addressed a problem that is relatively ill-posed. Specifically, there is not universal agreement about what constitutes a salient object when multiple observers are queried. This implies that some objects are more likely to be judged salient than others, and implies a relative rank exists on salient objects. The solution presented in this paper solves this more general problem that considers relative rank, and we propose data and metrics suitable to measuring success in a relative object saliency landscape. A novel deep learning solution is proposed based on a hierarchical representation of relative saliency and stage-wise refinement. We also show that the problem of salient object subitizing can be addressed with the same network, and our approach exceeds performance of any prior work across all metrics considered (both traditional and newly proposed).*

## 1. Introduction

The majority of work in salient object detection considers either a single salient object [37, 38, 7, 8, 31, 32, 9, 19, 17, 24, 39, 18] or multiple salient objects [13, 27, 36], but does not consider that what is salient may vary from one person to another, and certain objects may be met with more universal agreement concerning their importance.

There is a paucity of data that includes salient objects that are hand-segmented by multiple observers. It is important to note that any labels provided by a small number of observers (including one) does not allow for discerning the relative importance of objects. Implicit assignment of relative salience based on gaze data [33] also presents difficulties, given a different cognitive process than a calculated decision that involves manual labeling [16]. Moreover, gaze data is relatively challenging to interpret given factors such as centre bias, visuomotor constraints, and other latent factors [2, 1].

Therefore, in this paper we consider the problem of salient object detection more broadly. This includes detection of all salient regions in an image, and accounting for inter-observer variability by assigning confidence to different salient regions. We augment the PASCAL-S dataset [23] via further processing to provide ground truth in a form that accounts for relative salience. Success is measured against other algorithms based on the rank order of salient objects relative to ground truth orderings in addition to traditional metrics. Recent efforts also consider the problem of *salient object subitizing*. It is our contention that this determination should be possible by a model that provides detection of salient objects (see Fig. 1). We also allow our network to subitize.

As a whole, our work generalizes the problem of salient object detection, we present a new model that provides predictions of salient objects according to the traditional form of this problem, multiple salient object detection and relative ranking, and subitizing. Our results show state-of-the-art performance for all problems considered.

## 2. Background

### 2.1. Salient Object Detection:

Convolutional Neural Networks (CNNs) have raised the bar in performance for many problems in computer vision including salient object detection. CNN based models are able to extract more representative and complex features than hand crafted features used in less contemporary work [21, 34, 15] which has promoted widespread adoption.

---

*Both authors contributed equally to this work.

Some CNN based methods exploit superpixel and object region proposals to achieve accurate salient object detection [9, 19, 17, 22, 39, 18]. Such methods follow a multi-branch architecture where a CNN is used to extract semantic information across different levels of abstraction to generate an initial saliency prediction. Subsequently, new branches are added to obtain superpixels or object region proposals, which are used to improve precision of the predictions.

As an alternative to superpixels and object region proposals, other methods [26, 8, 37] predict saliency per-pixel by aggregating multi-level features. Luo et al. [26] integrate local and global features through a CNN that is structured as a multi-resolution grid. Hou et al. [8] implement stage-wise short connections between shallow and deeper feature maps for more precise detection and inferred the final saliency map considering only middle layer features. Zhang et al. [37] combine multi-level features as cues to generate and recursively fine-tune multi-resolution saliency maps which are refined by boundary preserving refinement blocks and then fused to produce final predictions.

Other methods [24, 31, 38] use an end-to-end encoder-decoder architecture that produces an initial coarse saliency map and then refines it stage-by-stage to provide better localization of salient objects. Liu and Han [24] propose a network that combines local contextual information step-by-step with a coarse saliency map. Wang et al. [31] propose a recurrent fully convolutional network for saliency detection that includes priors to correct initial saliency detection errors. Zhang et al. [38] incorporate a reformulated dropout after specific convolutional layers to quantify uncertainty in the convolutional features, and a new upsampling method to reduce artifacts of deconvolution which results in a better boundary for salient object detection.

In contrast to the above described approaches, we achieve spatial precision through stage-wise refinement by applying novel mechanisms to control information flow through the network while also importantly including a *stacking* strategy that implicitly carries the information necessary to determine relative saliency.

## 2.2. Salient Object Subitizing:

Recent work [35, 7] has also addressed the problem of subitizing salient objects in images. This task involves counting the number of salient objects, regardless of their importance or semantic category. The first salient object subitizing network proposed in [35] applies a feed-forward CNN to treat the problem as a classification task. He et al. [7] combine the subitizing task with detection by exploring the interaction between numeric and spatial representations. Our proposal provides a specific determination of the number of salient objects, recognizes variability in this number, and also provides output as a distribution that reflects this variability.

## 3. Proposed Network Architecture

We propose a new end-to-end framework for solving the problem of detecting multiple salient objects and ranking the objects according to their degree of salience. Our proposed salient object detection network is inspired by the success of convolution-deconvolution pipelines [28, 24, 12] that include a feed-forward network for initial coarse-level prediction. Then, we provide a stage-wise refinement mechanism over which predictions of finer structures are gradually restored. Fig. 2 shows the overall architecture of our proposed network. The encoder stage serves as a feature extractor that transforms the input image to a rich feature representation, while the refinement stages attempt to recover lost contextual information to yield accurate predictions and ranking.

We begin by describing how the initial coarse saliency map is generated in section 3.1. This is followed by a detailed description of the stage-wise refinement network, and multi-stage saliency map fusion in sections 3.2 and section 3.3 respectively.

### 3.1. Feed-forward Network for Coarse Prediction

Recent feed-forward deep learning models applied to high-level vision tasks (e.g. image classification [6, 30], object detection [29]) employ a cascade comprised of repeated convolution stages followed by spatial pooling. Down-sampling by pooling allows the model to achieve a highly detailed semantic feature representation with relatively poor spatial resolution at the deepest stage of encoding, also marked by spatial coverage of filters that is much larger in extent. The loss of spatial resolution is not problematic for recognition problems; however, pixel-wise labeling tasks (e.g. semantic segmentation, salient object detection) require pixel-precise information to produce accurate predictions. Thus, we choose Resnet-101 [6] as our encoder network (fundamental building block) due to its superior performance in classification and segmentation tasks. Following prior works on pixel-wise labeling [3, 12], we use the dilated ResNet-101 [3] to balance the semantic context and fine details, resulting in an output feature map reduced by a factor of 8. More specifically, given an input image $I \in \mathbb{R}^{h \times w \times d}$, our encoder network produces a feature map of size $\left\lfloor \frac{h}{8}, \frac{w}{8} \right\rfloor$. To augment the backbone of the encoder network with a top-down refinement network, we first attach one extra convolution layer with $3 \times 3$ kernel and 12 channels to obtain a *Nested Relative Salience Stack* (NRSS). Then, we append a *Stacked Convolutional Module* (SCM) to compute the coarse level saliency score for each pixel. It is worth noting that our encoder network is flexible enough to be replaced with any other baseline network e.g. VGG-16 [30], DenseNet-

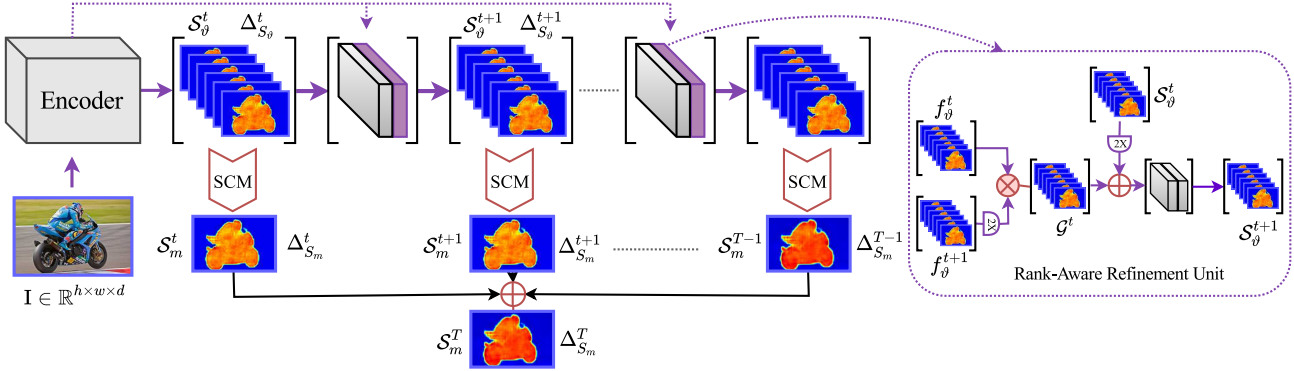

Figure 2. Illustration of our proposed network architecture. In the encoder network, the input image is processed with a feed-forward encoder to generate a coarse nested relative salience stack ($\mathcal{S}_\vartheta^t$). We append a Stacked Convolutional Module (SCM) on top of $\mathcal{S}_\vartheta^t$ to obtain a coarse saliency map $\mathcal{S}_m^t$. Then, a stage-wise refinement network, comprised of rank-aware refinement units (dotted box in the figure), successively refines each preceding NRSS ($\mathcal{S}_\vartheta^t$) and produces a refined NRSS ($\mathcal{S}_\vartheta^{t+1}$). A fusion layer combines predictions from all stages to generate the final saliency map ($\mathcal{S}_m^T$). We provide supervision ($\Delta_{S_\vartheta}^t, \Delta_{S_m}^t$) at the outputs ($\mathcal{S}_\vartheta^t, \mathcal{S}_m^t$) of each refinement stage. The architecture based on iterative refinement of a stacked representation is capable of effectively detecting multiple salient objects.

101 [10]. Moreover, we utilize atrous pyramid pooling [3] to gather more global contextual information. The described operations can be expressed as

$$\mathcal{S}_\vartheta^t = \mathcal{C}_{3\times3}(\mathcal{F}_s(I; \mathcal{W}); \Theta), \quad \mathcal{S}_m^t = \xi(\mathcal{S}_\vartheta^t) \qquad (1)$$

where $I$ is the input image and ($\mathcal{W}, \Theta$) denote the parameters of the convolution $\mathcal{C}$. $\mathcal{S}_\vartheta^t$ is the coarse level NRSS for stage $t$ that encapsulates different degrees of saliency for each pixel (akin to a prediction of the proportion of observers that might agree an object is salient), $\mathcal{S}_m^t$ refers to the coarse level saliency map, and $\xi$ refers to SCM. $\mathcal{F}_s(.)$ denotes the output feature map generated by the encoder network. The SCM consists of three convolutional layers for generating the desired saliency map. The initial convolutional layer has 6 channels with a $3\times3$ kernel, followed by two convolutional layers having 3 channels with $3 \times 3$ kernel and one channel with $1 \times 1$ kernel respectively. Each of the channels in the SCM learns a soft weight for each spatial location of the nested relative salience stack in order to label pixels based on confidence that they belong to a salient object.

## 3.2. Stage-wise Refinement Network

Most existing works [24, 32, 37, 8] that have shown success for salient object detection typically share a common structure of stage-wise decoding to recover per-pixel categorization. Although the deepest stage of an encoder has the richest possible feature representation, relying only on convolution and unpooling at the decoding stages to recover lost information may degrade the quality of predictions [12]. So, the spatial resolution that is lost at the deepest layer may be gradually recovered from earlier representations. This intuition appears in proposed refinement based models that include skip connections [25, 12, 37, 8] between encoder and decoder layers. However, how to effectively combine local and global contextual information remains an area deserving further analysis. Inspired by the success of refinement based approaches [25, 11, 12], we propose a multi-stage fusion based refinement network to recover lost contextual information in the decoding stage by combining an initial coarse representation with finer features represented at earlier layers. The refinement network is comprised of successive stages of rank-aware refinement units that attempt to recover missing spatial details in each stage of refinement and also preserve the relative rank order of salient objects. Each stage refinement unit takes the preceding NRSS with earlier finer scale representations as inputs and carries out a sequence of operations to generate a refined NRSS that contributes to obtain a refined saliency map. Note that refining the hierarchical NRSS implies that the refinement unit is leveraging the degree of agreement at different levels of SCMs to iteratively improve confidence in relative rank and overall saliency. As a final stage, refined saliency maps generated by the SCMs are fused to obtain the overall saliency map.

### 3.2.1 Rank-Aware Refinement Unit

Previous saliency detection networks [32, 24] proposed refinement across different levels by directly integrating representations from earlier features. Following [12], we integrate gate units in our rank-aware refinement unit that control the information passed forward to filter out the ambiguity relating to figure-ground and salient objects. The initial NRSS ($\mathcal{S}_\vartheta^t$) generated by the feed-forward encoder provides input for the first refinement unit. Note that one can interpret $\mathcal{S}_\vartheta^t$ as the predicted saliency map in the decoding process, but our model forces the channel dimension to be the same as the number of participants involved in labeling salient objects. The refinement unit takes the gated feature map $\mathcal{G}^t$ generated by the gate unit [12] as a second input. As suggested by [12], we obtain $\mathcal{G}^t$ by combining two consecutive feature maps ($f_\vartheta^t$ and $f_\vartheta^{t+1}$) from the encoder network (see dotted box in Fig. 2). We first upsample the preceding $\mathcal{S}_\vartheta^t$ to double its size. A transformation function $\mathcal{T}_f$ comprised of a sequence of operations is applied on upsampled $\mathcal{S}_\vartheta^t$ and $\mathcal{G}^t$ to obtain the refined NRSS ($\mathcal{S}_\vartheta^{t+1}$). We then append the *SCM* module on top of $\mathcal{S}_\vartheta^{t+1}$ to generate the refined saliency map $\mathcal{S}_m^{t+1}$. Finally, the predicted $\mathcal{S}_\vartheta^{t+1}$ is fed to the next stage rank-aware refinement unit. Note that, we only forward the NRSS to the next stage, allowing the network to learn contrast between different levels of confidence for salient objects. Unlike other approaches, we apply supervision for both of the refined NRSS and the refined saliency map. The procedure for obtaining the refined NRSS and the refined saliency map for all stages is identical. The described operations may be summarized as follows:

$$\mathcal{S}_\vartheta^{t+1} = w^b * \mathcal{T}_f(\mathcal{G}^t, u(\mathcal{S}_\vartheta^t)), \ S_m^{t+1} = w_s^b * \xi(\mathcal{S}_\vartheta^{t+1}) \quad (2)$$

where $u$ represents the upsample operation; $w^b$ and $w_s^b$ denotes the parameter for the transformation function $\mathcal{T}_f$ and SCM ($\xi$ in the equation) respectively. Note that $t$ refers to particular stage of the refinement process.

### 3.3. Multi-Stage Saliency Map Fusion

Predicted saliency maps at different stages of the refinement units are capable of finding the location of salient regions with increasingly sharper boundaries. Since all the rank-aware refinement units are stacked together on top of each other, the network allows each stage to learn specific features that are of value in the refinement process. These phenomena motivate us to combine different level SCMs predictions, since the internal connection between them is not explicitly present in the network structure. To facilitate interaction, we add a fusion layer at the end of network that concatenates the predicted saliency maps of different stages, resulting in a fused feature map $\mathcal{S}_m^{\hat{f}}$. Then, we apply a $1 \times 1$ convolution layer $\Upsilon$ to produce the final predicted saliency map $\mathcal{S}_m^T$ of our network. Note that our network has T predictions, including one fused prediction and T-1 stage-wise predictions. We can write the operations as follows:

$$\mathcal{S}_m^{\hat{f}} = \breve{\eth}(\mathcal{S}_m^t, \mathcal{S}_m^{t+1}, ...., \mathcal{S}_m^{T-1}), \ \mathcal{S}_m^T = w^f * \Upsilon(\mathcal{S}_m^{\hat{f}}) \quad (3)$$

where $\breve{\eth}$ denotes the cross channel concatenation; $w^f$ is the resultant parameter for obtaining the final prediction.

### 3.4. Stacked Representation of Ground-truth

The ground-truth for salient object detection or segmentation contains a set of numbers defining the degree of saliency for each pixel. The traditional way of generating binary masks is by thresholding which implies that there is no notion of relative salience. Since we aim to explicitly model observer agreement, using traditional binary ground-truth masks is unsuitable. To address this problem, we propose to generate a set of stacked ground-truth maps that corresponds to different levels of saliency (defined by inter-observer agreement). Given a ground-truth saliency map $\mathcal{G}_m$, we obtain a stack $\mathcal{G}_\vartheta$ of $N$ ground-truth maps ($\mathcal{G}_i, \mathcal{G}_{i+1}, ....., \mathcal{G}_N$) where each map $\mathcal{G}_i$ includes a binary indication that at least $i$ observers judged an object to be salient (represented at a per-pixel level). $N$ is the number of different participants involved in labeling the salient objects. The stacked ground-truth saliency maps $\mathcal{G}_\vartheta$ provides better separation for multiple salient objects (see Eq. (4) for illustration) and also naturally acts as the relative rank order that allows the network to learn to focus on degree of salience. It is important to note the nested nature of the stacked ground truth wherein $\mathcal{G}_{i+1} \subseteq \mathcal{G}_i$. This is important conceptually as a representation wherein $\mathcal{G}_i = 1 \iff$ exactly $i$ observers agree, results in zeroed layers in the ground truth stack, and large changes to ground truth based on small differences in degree of agreement.

$$\mathcal{G}_\vartheta = \left[\begin{array}{c}\mathcal{G}_i\end{array}\right] \left[\begin{array}{c}\mathcal{G}_{i+1}\end{array}\right] \left[\begin{array}{c}\mathcal{G}_{i+2}\end{array}\right] \left[...\right] \left[\begin{array}{c}\mathcal{G}_N\end{array}\right] \quad (4)$$

### 3.5. Salient Object Subitizing Network

Previous works [35, 7] treat subitizing as a straightforward classification task. Similar to our multiple salient object detection network, the subitizing network is also based on ResNet-101 [6] except we remove the last block. We append a fully connected layer at the end to generate confidence scores for each of 0, 1, 2, 3, and 4+ salient objects existing in the input image followed by another fully connected layer leads to generate final confidence scores for each category. The reasoning behind this is that a single layer allows for accumulation of confidence tied to salience while two layers allows for reasoning about relative salience. We use our pre-trained detection model to train the subitizing network. As a classifier, the subitizing network reduces two cross entropy losses $\ell_{sub}^1(c, n)$ and $\ell_{sub}^{\hat{f}}(c_f, n)$ between the number of salient objects $n$ in ground-truth, and the total predicted objects.

**A New Dataset for Salient Object Subitizing:** Since salient object subitizing is not a widely addressed problem, a limited number of datasets [35] were created. In order to facilitate the study of this problem in more complex scenarios, we create the subitizing ground-truth for the Pascal-S dataset [23] that provides instance-wise counting as labels. The distribution of the images in Pascal-S dataset with respect to different categories is shown in Table 1. It is an evident from the table that, there is a considerable number of images with more than two salient objects but only few images with more than 7. We initially include all instances of salient objects in the labeling process. To reduce imbalance between different categories, we create another ground-truth set where we only categorize the images as 1, 2, 3, and 4+ salient objects.

| # Salient Object | 1 | 2 | 3 | 4 | 5 | 6 | 7 | 8+ | Total |
|---|---|---|---|---|---|---|---|---|---|
| #Images | 300 | 227 | 136 | 72 | 43 | 28 | 18 | 26 | 850 |
| Distribution (%) | 0.35 | 0.27 | 0.16 | 0.08 | 0.05 | 0.03 | 0.02 | 0.03 | 1 |

Table 1. Count and percentage of images corresponding to different numbers of salient objects in the Pascal-S dataset.

### 3.6. Training the Network

Our proposed network produces a sequence of nested relative salience stacks (NRSS) and saliency maps at each stage of refinement; however, we are principally interested in the final fused saliency map. Each stage of the network is encouraged to repeatedly produce NRSS and a saliency map with increasingly finer details by leveraging preceding NRSS representations. We apply an auxiliary loss at the output of each refinement stage along with an overall master loss at the end of the network. Both of the losses help the optimization process. In more specific terms, let $I \in \mathbb{R}^{h \times w \times 3}$ be a training image with ground-truth saliency map $\mathcal{G}_m \in \mathbb{R}^{h \times w}$. As described in section 3.4, we generate a stack of ground-truth saliency maps $\mathcal{G}_\vartheta \in \mathbb{R}^{h \times w \times 12}$. To apply supervision on the NRSS ($S_\vartheta^t$) and saliency map $S_m^t$, we first down-sample $\mathcal{G}_\vartheta$ and $\mathcal{G}_m$ to the size of $S_\vartheta^t$ generated at each stage resulting in $\mathcal{G}_\vartheta^t$ and $\mathcal{G}_m^t$. Then, at each refinement stage we define pixel-wise euclidean loss $\Delta_{S_\vartheta}^t$ and $\Delta_{S_m}^t$ to measure the difference between $(S_\vartheta^t, \mathcal{G}_\vartheta^t)$ and $(S_m^t, \mathcal{G}_m^t)$ respectively. We can summarize these operations as:

$$\Delta_{S_\vartheta}^t(W) = \frac{1}{2dN} \sum_{i=1}^{d} \sum_{z=1}^{N} (x_i(z) - y_i(z))^2$$

$$\Delta_{S_m}^t(W) = \frac{1}{2d} \sum_{i=1}^{d} (x_i - y_i)^2$$

$$L_{aux}^t(W) = \Delta_{S_\vartheta}^t + \Delta_{S_m}^t \quad (5)$$

where $x \in \mathbb{R}^d$ and $y \in \mathbb{R}^d$ ($d$ denotes the spatial resolution) are the vectorized ground-truth and predicted saliency map. $x_i$ and $y_i$ refer to a particular pixel of $S_\vartheta^t$ and $G_\vartheta^t$ respectively. $W$ denotes the parameters of whole network and $N$ refers to total number of ground-truth slices (N =12 in our case). The final loss function of the network combining master and auxiliary losses can be written as:

$$L_{final}(W) = L_{mas}(W) + \sum_{t=1}^{T-1} \lambda_t L_{aux}^t(W) \quad (6)$$

where $L_{mas}(W)$ refers to the euclidean loss function computed on the final predicted saliency map $\mathcal{S}_m^T$. We set $\lambda_t$ to 1 for all stages to balance the loss, which remains continuously differentiable. Each stage of prediction contains information related to two predictions, allowing our network to propagate supervised information from deep layers. This also begins with aligning the weights with the initial coarse representation, leading to a coarse-to-fine learning process. The fused prediction generally appears much better than other stage-wise predictions since it contains the aggregated information from all the refinement stages. For saliency inference, we can simply feed an image of arbitrary size to the network and use the fused prediction as our final saliency map.

## 4. Experiments

The core of our model follows a structure based on ResNet-101 [6] with pre-trained weights to initialize the encoder portion. A few variants of the basic architecture are proposed, and we report numbers for the following variants that are described in what follows:

**RSDNet:** This network includes dilated ResNet-101 [3] + NRSS + SCM. **RSDNet-A:** This network is the same as RSDNet except the ground-truth is scaled by a factor of 1000, encouraging the network to explicitly learn deeper contrast. **RSDNet-B:** The structure follows RSDNet except that an atrous pyramid pooling module is added. **RSDNet-C:** RSDNet-B + the ground-truth scaling. **RSDNet-R:** RSDNet with stage-wise rank-aware refinement units + multi-stage saliency map fusion.

### 4.1. Datasets and Evaluation Metrics

**Datasets:** The Pascal-S dataset includes 850 natural images with multiple complex objects derived from the PASCAL VOC 2012 validation set [4]. We randomly split the Pascal-S dataset into two subsets (425 for training and 425 for testing). In this dataset, salient object labels are based on an experiment using 12 participants to label salient objects. Virtually all existing approaches for salient object segmentation or detection threshold the ground-truth saliency map to obtain a binary saliency map. This operation seems somewhat arbitrary since the threshold can require consensus among $k$ observers, and the value of $k$ varies from one study to another. This is one of the most highly used salient object segmentation datasets, but is unique in having multiple explicitly tagged salient regions provided by a reasonable sample size of observers. Since a key objective of this work is to rank salient objects in an image, we use the original ground-truth maps (each pixel having a value corresponding to the number of observers that deemed it to be a salient object) rather than trying to predict a binary output based on an arguably contentious thresholding process.

**Evaluation Metrics:** For the multiple salient object detection task, we use four different standard metrics to measure performance including precision-recall (PR) curves, F-measure (maximal along the curve), Area under ROC curve (AUC), and mean absolute error (MAE). Since some of these rely on binary decisions, we threshold the ground-truth saliency map based on the number of participants that deem an object salient, resulting in 12 binary ground truth maps. For each binary ground truth map, multiple thresholds of a predicted saliency map allow for calculation of the true positive rate (TPR), false positive rate (FPR), precision and recall, and corresponding ROC and PR curves. Given that methods that predate this work are trained based on varying thresholds and consider a binary ground truth map, scores are reported based on the binary ground truth map that produces the best AUC or F-measure score (and the corresponding curves are shown). Max F-measure, average F-measure and median F-measure are also reported to provide a sense of how performance varies as a function of the threshold chosen. We also report the MAE score i.e. the average pixel-wise difference between the predicted saliency map and the binary ground-truth map that produces the minimum score.

In ordered to evaluate the rank order of salient objects, we introduce the *Salient Object Ranking* (SOR) metric which is defined as the Spearman's Rank-Order Correlation between the ground truth rank order and the predicted rank order of salient objects. SOR score is normalized to [0 1] for ease of interpretation. Scores are reported based on the average SOR score for each method considering the whole dataset.

### 4.2. Performance Comparison with State-of-the-art

The problem of evaluating salient detection models is challenging in itself which has contributed to differences among benchmarks that are used. In light of these considerations, the specific evaluation we have applied to all the methods aims to remove any advantages of one algorithm over another. We compare our proposed method with recent state-of-the-art approaches, including Amulet [37], UCF [38], DSS [8], NLDF [26], DHSNet [24], MDF [18], ELD [17], MTDS [22], MC [39], HS [34], HDCT [15], DSR [21], and DRFI [14]. For fair comparison, we build the evaluation code based on the publicly available code provided in [20] and we use saliency maps provided by authors of models compared against, or by running their pre-trained models with recommended parameter settings.

**Quantitative Evaluation:** Table 2 shows the performance score of all the variants of our model, and other recent methods on salient object detection. It is evident that, RSDNet-R outperforms other recent approaches for all evaluation metrics, which establishes the effectiveness of our proposed hierarchical nested relative salience stack. From the results we have few fundamental observations: (1) Our network improves the max F-measure by a considerable margin on the Pascal-S dataset which indicates that our model is general enough that it achieves higher precision with higher recall (see Fig. 3). (2) Our model decreases the overall MAE on the Pascal-S dataset and achieves higher area under the ROC curve (AUC) score compared to the baselines shown in Fig. 3. (3) Although our model is only trained on a subset of Pascal-S, it significantly outperforms other algorithms that also leverage large-scale saliency datasets. Overall, this analysis hints at strengths of the proposed hierarchical stacked refinement strategy to provide a more accurate saliency map. In addition, it is worth mentioning that RDSNet-R outperforms all the recent deep learning based methods intended for salient object detection/segmentation without any post-processing techniques such as CRF that are typically used to boost scores.

| * | AUC | max-$F_m$ | med-$F_m$ | avg-$F_m$ | MAE | **SOR** |
|---|---|---|---|---|---|---|
| DRFI [14] | 0.887 | 0.716 | 0.583 | 0.504 | 0.216 | 0.726 |
| DSR [21] | 0.871 | 0.696 | 0.628 | 0.583 | 0.186 | 0.728 |
| HDCT [15] | 0.809 | 0.654 | 0.567 | 0.523 | 0.214 | 0.645 |
| HS [34] | 0.837 | 0.702 | 0.634 | 0.596 | 0.263 | 0.714 |
| MC [39] | 0.870 | 0.717 | 0.616 | 0.573 | 0.216 | 0.732 |
| MTDS [22] | 0.941 | 0.805 | 0.731 | 0.664 | 0.176 | 0.782 |
| ELD [17] | 0.916 | 0.789 | 0.784 | 0.774 | 0.123 | 0.792 |
| MDF [18] | 0.892 | 0.787 | 0.746 | 0.730 | 0.138 | 0.768 |
| DHSNet [24] | 0.927 | 0.837 | 0.833 | 0.822 | 0.092 | 0.781 |
| NLDF [26] | 0.933 | 0.846 | 0.843 | 0.836 | 0.099 | 0.783 |
| DSS [8] | 0.918 | 0.841 | 0.838 | 0.830 | 0.099 | 0.770 |
| AMULET [37] | 0.957 | 0.865 | 0.854 | 0.841 | 0.097 | 0.788 |
| UCF [38] | 0.959 | 0.858 | 0.840 | 0.813 | 0.123 | 0.792 |
| **RSDNet** | 0.972 | 0.873 | 0.854 | 0.834 | 0.091 | 0.825 |
| RSDNet-**A** | 0.973 | 0.874 | 0.851 | 0.796 | 0.103 | 0.838 |
| RSDNet-**B** | 0.969 | 0.877 | 0.857 | 0.831 | 0.100 | 0.840 |
| RSDNet-**C** | 0.972 | 0.874 | 0.850 | 0.795 | 0.110 | 0.848 |
| RSDNet-**R** | 0.971 | 0.880 | 0.861 | 0.837 | 0.090 | 0.852 |

Table 2. Quantitative comparison of methods including AUC, max F-measure (higher is better), median F-measure, average F-measure, MAE (lower is better), and SOR (higher is better). The best three results are shown in red, violet and blue respectively.

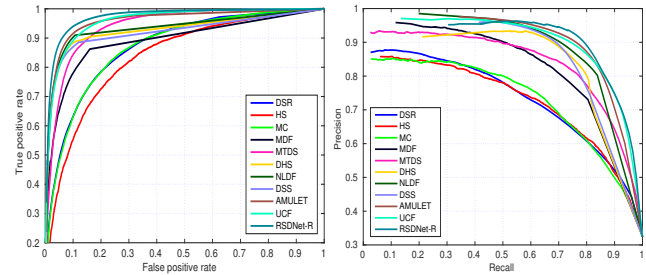

Figure 3. Left: ROC curves corresponding to different state-of-the-art methods. Right: Precision-Recall curves for salient object detection corresponding to a variety of algorithms.

**Qualitative Evaluation:** Fig. 4 depicts a visual comparison of RSDNet-R with respect to other state-of-the-art methods. We can see that our method can predict salient regions accurately and produces output closer to ground-truth maps in various challenging cases e.g., instances touching the image boundary (1$^{st}$ & 2$^{nd}$ rows), multiple instances of same object (3$^{rd}$ row). The nested relative salience stack at each stage provides distinct representations to differentiate between multiple salient objects and allows for reasoning about their relative salience to take place.

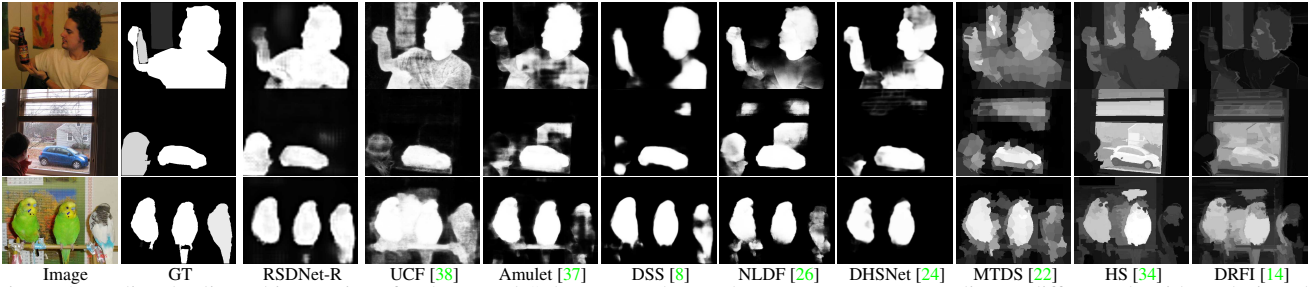

Figure 4. Predicted salient object regions for the Pascal-S dataset. Each row shows outputs corresponding to different algorithms designed for the salient object detection/segmentation task.

### 4.2.1 Application: Ranking by Detection

As salient instance ranking is a completely new problem, there is not existing benchmark. In order to promote this direction of studying this problem, we are interested in finding the ranking of salient objects from the predicted saliency map. Rank order of a salient instance is obtained by averaging the degree of saliency within that instance mask.

$$\text{Rank}(\mathcal{S}_m^T(\delta)) = \frac{\sum_{i=1}^{\rho_\delta} \delta(x_i, y_i)}{\rho_\delta} \quad (7)$$

where $\delta$ represents a particular instance of the predicted saliency map ($\mathcal{S}_m^T$), $\rho_\delta$ denotes total numbers of pixels $\delta$ contains, and $\delta(x_i, y_i)$ refers to saliency score for the pixel $(x_i, y_i)$. While there may exist alternatives for defining rank order, this is an intuitive way of assigning this score. With that said, we expect that this is another interesting nuance of the problem to explore further; specifically salience vs. scale, and part-whole relationships. Note that we do not need to change the network architecture to obtain the desired ranking. Instead we use the provided instance-wise segmentation and saliency map to calculate the ranking for each image.

To demonstrate the effectiveness of our approach, we compare the overall ranking score with recent state-of-the-art approaches. It is worth noting that no prior methods report results for salient instance ranking. We apply the proposed SOR evaluation metric to report how different models gauge relative salience. The last column in Table 2 shows the SOR score of our approach and comparisons with other state-of-the-art methods. We achieve 85.2% correlation score for the best variant of our model. The proposed method significantly outperforms other approaches in ranking multiple salient objects and our analysis shows that learning salient object detection implicitly learns rank to some extent, but explicitly learning rank can also improve salient object detection irrespective of how the ground truth is defined. Fig. 5 shows a qualitative comparison of the state-of-the-art approaches designed for salient object detection. Note that the role of ranking for more than three objects is particularly pronounced.

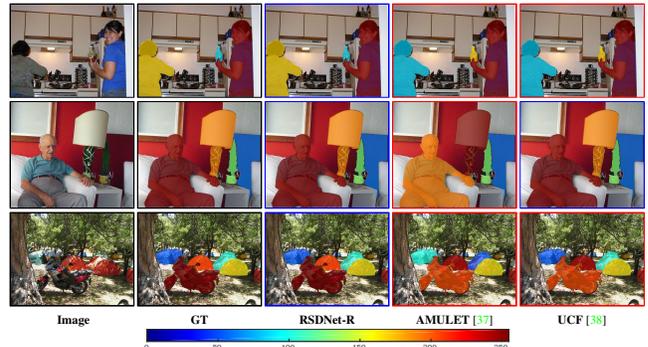

Figure 5. Qualitative depiction of rank order of salient objects. Relative rank is indicated by the assigned color. Blue and red image borders indicate correct and incorrect ranking respectively.

### 4.2.2 Application: Salient Object Subitizing

As mentioned prior, salient object detection, ranking, and subitizing are interrelated. It is therefore natural to consider whether salient region prediction and ranking provide guidance to subitize. A copy of the detection network is further trained to perform subitizing on Pascal-S. For simplicity (and in line with prior work [35, 7]), we train our system only for predicting objects either for 1, 2, 3, or 4+ and report the Average Precision (AP) [5] in Table 3. Since this is the first work to perform subitizing on the Pascal-S dataset, we do not have any baselines to compare with. To make comparison possible, we fine-tune and evaluate our model on the SOS dataset [35], and report the AP and weighted AP (overall) scores in Table 4. Our proposed model achieves state-of-the-art results on this dataset compared to baselines.

| * | 1 | 2 | 3 | 4+ | mean |
|---|---|---|---|---|---|
| RSDNet | 0.62 | 0.42 | 0.20 | 0.55 | 0.45 |

Table 3. Average Precision (AP) on Pascal-S dataset.

| * | 0 | 1 | 2 | 3 | 4+ | mean | overall |
|---|---|---|---|---|---|---|---|
| count | 338 | 617 | 219 | 137 | 69 | - | - |
| % | 0.24 | 0.45 | 0.16 | 0.10 | 0.05 | - | - |
| CNN [35] | 0.92 | 0.82 | 0.34 | 0.31 | 0.56 | 0.59 | 0.70 |
| SOS [35] | 0.93 | 0.90 | 0.51 | 0.48 | 0.65 | 0.69 | 0.79 |
| **RSDNet** | **0.95** | **0.92** | **0.61** | **0.59** | **0.67** | **0.75** | **0.83** |

Table 4. Overall and Average Precision (AP) on the SOS dataset.

| * | S-1 | | S-2 | | S-3 | | S-4 | | S-5 | | S-6 | | S-7 | | S-8 | | S-9 | | S-10 | | S-11 | | S-12 | |
|---|---|---|---|---|---|---|---|---|---|---|---|---|---|---|---|---|---|---|---|---|---|---|---|---|
| | AUC | Fm | AUC | Fm | AUC | Fm | AUC | Fm | AUC | Fm | AUC | Fm | AUC | Fm | AUC | Fm | AUC | Fm | AUC | Fm | AUC | Fm | AUC | Fm |
| NLDF [26] | 0.900 | 0.846 | 0.922 | 0.840 | 0.931 | 0.836 | 0.933 | 0.831 | 0.930 | 0.827 | 0.922 | 0.821 | 0.925 | 0.818 | 0.913 | 0.809 | 0.897 | 0.802 | 0.865 | 0.782 | 0.812 | 0.751 | 0.660 | 0.680 |
| DSS [8] | 0.883 | 0.841 | 0.906 | 0.839 | 0.916 | 0.832 | 0.918 | 0.825 | 0.915 | 0.821 | 0.910 | 0.819 | 0.912 | 0.816 | 0.901 | 0.805 | 0.886 | 0.799 | 0.855 | 0.779 | 0.802 | 0.745 | 0.651 | 0.675 |
| AMULET [37] | 0.932 | 0.865 | 0.949 | 0.856 | 0.954 | 0.850 | 0.957 | 0.847 | 0.952 | 0.840 | 0.944 | 0.834 | 0.946 | 0.829 | 0.933 | 0.819 | 0.918 | 0.813 | 0.884 | 0.791 | 0.827 | 0.760 | 0.671 | 0.688 |
| UCF [38] | 0.940 | 0.858 | 0.955 | 0.845 | 0.959 | 0.838 | 0.959 | 0.831 | 0.956 | 0.829 | 0.947 | 0.825 | 0.949 | 0.823 | 0.935 | 0.813 | 0.918 | 0.806 | 0.885 | 0.785 | 0.827 | 0.754 | 0.672 | 0.689 |
| **RSDNet** | 0.950 | 0.872 | 0.966 | 0.873 | 0.970 | 0.870 | **0.972** | 0.868 | **0.967** | 0.860 | **0.957** | 0.854 | **0.957** | 0.850 | **0.945** | 0.842 | **0.926** | 0.834 | **0.893** | 0.812 | 0.836 | 0.774 | 0.676 | **0.705** |
| RSDNet-**A** | **0.952** | 0.874 | **0.967** | 0.874 | **0.972** | 0.871 | **0.973** | **0.869** | **0.968** | 0.860 | **0.958** | **0.856** | **0.958** | **0.853** | **0.946** | **0.846** | **0.928** | **0.836** | **0.895** | **0.815** | **0.837** | **0.778** | **0.677** | **0.707** |
| RSDNet-**B** | 0.948 | **0.877** | 0.963 | **0.877** | 0.968 | **0.873** | 0.969 | **0.871** | 0.964 | **0.862** | 0.954 | **0.856** | 0.954 | 0.852 | 0.942 | 0.844 | 0.923 | 0.833 | 0.889 | 0.810 | 0.831 | 0.774 | 0.672 | 0.702 |
| RSDNet-**C** | **0.955** | 0.874 | **0.968** | 0.872 | **0.971** | 0.869 | **0.972** | 0.867 | **0.967** | 0.859 | **0.958** | 0.851 | **0.958** | 0.851 | **0.946** | 0.843 | **0.928** | 0.835 | **0.895** | 0.813 | **0.838** | 0.775 | **0.678** | 0.699 |
| RSDNet-**R** | 0.951 | **0.880** | 0.965 | **0.879** | 0.969 | **0.874** | 0.971 | **0.871** | 0.966 | **0.866** | 0.956 | **0.859** | 0.956 | **0.854** | 0.944 | **0.849** | 0.925 | **0.838** | 0.892 | **0.815** | 0.833 | **0.776** | 0.674 | 0.701 |

Table 5. Quantitative comparison (AUC & Fm) with state-of-the-art methods across all ground truth thresholds, each corresponding to agreement among a specific number participants. Best and second best scores are shown in red and blue respectively.

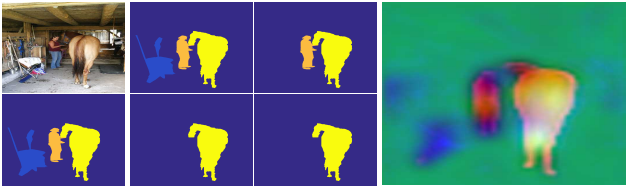

Figure 6. Visualization of Principal component analysis (PCA) for the final prediction stack (NRSS) of our model. The first column shows the image and its ground truth. Second and third columns show a selection of ground truth stack slices. The final column provides a visualization of the top three principal components for our predicted stack as an RGB image. Note that the contribution of the top three components itself is diagnostic with respect to relative salience.

### 4.3. Examining the Nested Relative Salience Stack

Comparison of slices of the nested relative salience stack can be challenging as differences between some layer pairs may be subtle, and contrast can differ across layers. We therefore examine variability among NRSS layers through Principal component analysis (PCA) to determine regions where greatest variability (and signal) exists. Fig. 6 shows the top three principal components as an RGB image where the first principal component (which captures the most variance across layers) is mapped to the R-channel, the second principal component is mapped to the G-channel and so forth. Salient areas in the ground truth are captured in the variability across layers demonstrating the value of our stacking mechanism for saliency ranking. Moreover, it is nearly possible to read a relative ranking directly from this visualization wherein high values for the first 2 eigenvectors result in yellow, the first only red, etc.

We also report the AUC score and max F-measure for each slice (denoted as S) in Table 5. Compared to baselines, our proposed method achieves better scores across all ground truth thresholds, that correspond to the different numbers of participants showing agreement that an object is salient. This further shows the effectiveness of the stacking mechanism and predicting relative salience, which results in improvements no matter how the ground truth is determined (if considered as a binary quantity).

### 4.4. Failure Cases

Despite good performance for the majority of cases; there are instances that are more challenging to predict (see Fig. 7). Sometimes, the ground truth has multiple objects with the same degree of saliency (ties in participants agreeing) (see 1st row in Fig. 7). Other failures of ranking happen when there is considerable diversity in agreement on what is salient in an image (as shown in the second row) or when there is occlusion among two objects which have a relatively close degree of saliency as shown in the last row.

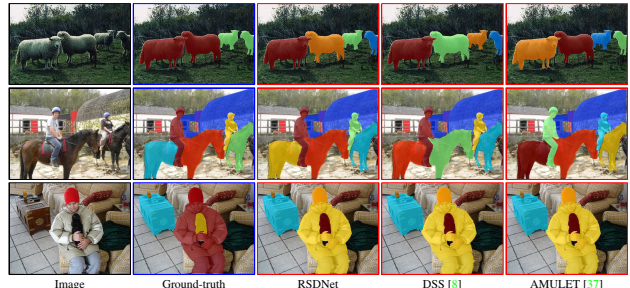

Figure 7. Shown are some illustrative examples of disagreements in rank between model and ground truth. These are most common for ties in the ground truth, and for scenes with many salient objects.

### 5. Conclusion

In this paper, we have presented a neural framework for detecting, ranking, and subitizing multiple salient objects that introduces a stack refinement mechanism to achieve better performance. Central to the success of this approach, is how to represent relative saliency both in terms of ground truth, and *in network* in a manner that produces stable performance. We highlight the fact that to date, salient object detection has assumed a relatively limited, and sometimes inconsistent problem definition. Comprehensive experiments demonstrate that the proposed architecture outperforms state-of-the-art approaches across a broad gamut of metrics.

**Acknowledgments:** The authors acknowledge financial support from the NSERC Canada Discovery Grants program, University of Manitoba GETS and UMGF funding, and the support of the NVIDIA Corporation GPU Grant Program.

# References

[1] A. Açık, A. Bartel, and P. Koenig. Real and implied motion at the center of gaze. *Journal of Vision*, 14(1):2–2, 2014. 1

[2] N. D. Bruce, C. Wloka, N. Frosst, S. Rahman, and J. K. Tsotsos. On computational modeling of visual saliency: Examining what's right, and what's left. *Vision research*, 116:95–112, 2015. 1

[3] L.-C. Chen, G. Papandreou, I. Kokkinos, K. Murphy, and A. L. Yuille. Deeplab: Semantic image segmentation with deep convolutional nets, atrous convolution, and fully connected crfs. *TPAMI*, 40(4):834–848, 2018. 2, 3, 5

[4] M. Everingham, L. Van Gool, C. K. Williams, J. Winn, and A. Zisserman. The pascal visual object classes (voc) challenge. *IJCV*, 88(2):303–338, 2010. 5

[5] M. Everingham, L. Van Gool, C. K. I. Williams, J. Winn, and A. Zisserman. The PASCAL Visual Object Classes Challenge 2007. 7

[6] K. He, X. Zhang, S. Ren, and J. Sun. Deep residual learning for image recognition. In *CVPR*, 2016. 2, 4, 5

[7] S. He, J. Jiao, X. Zhang, G. Han, and R. W. Lau. Delving into salient object subitizing and detection. In *CVPR*, 2017. 1, 2, 4, 7

[8] Q. Hou, M.-M. Cheng, X. Hu, A. Borji, Z. Tu, and P. Torr. Deeply supervised salient object detection with short connections. In *CVPR*, 2017. 1, 2, 3, 6, 7, 8

[9] P. Hu, B. Shuai, J. Liu, and G. Wang. Deep level sets for salient object detection. In *CVPR*, 2017. 1, 2

[10] G. Huang, Z. Liu, L. Maaten, and K. Q. Weinberger. Densely connected convolutional networks. In *CVPR*, 2017. 3

[11] M. A. Islam, S. Naha, M. Rochan, N. Bruce, and Y. Wang. Label refinement network for coarse-to-fine semantic segmentation. *arXiv:1703.00551*, 2017. 3

[12] M. A. Islam, M. Rochan, N. D. Bruce, and Y. Wang. Gated feedback refinement network for dense image labeling. In *CVPR*, 2017. 2, 3, 4

[13] S. Jia, Y. Liang, X. Chen, Y. Gu, J. Yang, N. Kasabov, and Y. Qiao. Adaptive location for multiple salient objects detection. In *NIPS*, 2015. 1

[14] H. Jiang, J. Wang, Z. Yuan, Y. Wu, N. Zheng, and S. Li. Salient object detection: A discriminative regional feature integration approach. In *CVPR*, 2013. 6, 7

[15] J. Kim, D. Han, Y.-W. Tai, and J. Kim. Salient region detection via high-dimensional color transform. In *CVPR*, 2014. 2, 6

[16] K. Koehler, F. Guo, S. Zhang, and M. P. Eckstein. What do saliency models predict? *Journal of vision*, 14(3):14–14, 2014. 1

[17] G. Lee, Y.-W. Tai, and J. Kim. Deep saliency with encoded low level distance map and high level features. In *CVPR*, 2016. 1, 2, 6

[18] G. Li and Y. Yu. Visual saliency based on multiscale deep features. In *CVPR*, 2015. 1, 2, 6

[19] G. Li and Y. Yu. Deep contrast learning for salient object detection. In *CVPR*, 2016. 1, 2

[20] X. Li, Y. Li, C. Shen, A. Dick, and A. Van Den Hengel. Contextual hypergraph modeling for salient object detection. In *ICCV*, 2013. 6

[21] X. Li, H. Lu, L. Zhang, X. Ruan, and M.-H. Yang. Saliency detection via dense and sparse reconstruction. In *ICCV*, 2013. 2, 6

[22] X. Li, L. Zhao, L. Wei, M.-H. Yang, F. Wu, Y. Zhuang, H. Ling, and J. Wang. Deepsaliency: Multi-task deep neural network model for salient object detection. *TIP*, 2016. 2, 6, 7

[23] Y. Li, X. Hou, C. Koch, J. M. Rehg, and A. L. Yuille. The secrets of salient object segmentation. In *CVPR*, 2014. 1, 4

[24] N. Liu and J. Han. Dhsnet: Deep hierarchical saliency network for salient object detection. In *CVPR*, 2016. 1, 2, 3, 6, 7

[25] J. Long, E. Shelhamer, and T. Darrell. Fully convolutional networks for semantic segmentation. In *CVPR*, 2015. 3

[26] Z. Luo, A. Mishra, A. Achkar, J. Eichel, S. Li, and P.-M. Jodoin. Non-local deep features for salient object detection. In *CVPR*, 2017. 2, 6, 7, 8

[27] M. Najibi, F. Yang, Q. Wang, and R. Piramuthu. Towards the success rate of one: Real-time unconstrained salient object detection. *arXiv:1708.00079*, 2017. 1

[28] H. Noh, S. Hong, and B. Han. Learning deconvolution network for semantic segmentation. In *ICCV*, 2015. 2

[29] S. Ren, K. He, R. Girshick, and J. Sun. Faster r-cnn: Towards real-time object detection with region proposal networks. In *NIPS*, 2015. 2

[30] K. Simonyan and A. Zisserman. Very deep convolutional networks for large-scale image recognition. *arXiv:1409.1556*, 2014. 2

[31] L. Wang, L. Wang, H. Lu, P. Zhang, and X. Ruan. Saliency detection with recurrent fully convolutional networks. In *ECCV*, 2016. 1, 2

[32] T. Wang, A. Borji, L. Zhang, P. Zhang, and H. Lu. A stagewise refinement model for detecting salient objects in images. In *CVPR*, 2017. 1, 3

[33] C. Xia, J. Li, X. Chen, A. Zheng, and Y. Zhang. What is and what is not a salient object? learning salient object detector by ensembling linear exemplar regressors. In *CVPR*, 2017. 1

[34] Q. Yan, L. Xu, J. Shi, and J. Jia. Hierarchical saliency detection. In *CVPR*, 2013. 2, 6, 7

[35] J. Zhang, S. Ma, M. Sameki, S. Sclaroff, M. Betke, Z. Lin, X. Shen, B. Price, and R. Mech. Salient object subitizing. In *CVPR*, 2015. 2, 4, 7

[36] J. Zhang, S. Sclaroff, Z. Lin, X. Shen, B. Price, and R. Mech. Unconstrained salient object detection via proposal subset optimization. In *CVPR*, 2016. 1

[37] P. Zhang, D. Wang, H. Lu, H. Wang, and X. Ruan. Amulet: Aggregating multi-level convolutional features for salient object detection. In *ICCV*, 2017. 1, 2, 3, 6, 7, 8

[38] P. Zhang, D. Wang, H. Lu, H. Wang, and B. Yin. Learning uncertain convolutional features for accurate saliency detection. In *ICCV*, 2017. 1, 2, 6, 7, 8

[39] R. Zhao, W. Ouyang, H. Li, and X. Wang. Saliency detection by multi-context deep learning. In *CVPR*, 2015. 1, 2, 6